\documentclass[sigconf]{acmart}
\usepackage{pifont}
\usepackage{latexsym}

\usepackage{balance}
\usepackage{helvet}  
\usepackage{courier}  
\usepackage{url}  
\usepackage{graphicx}  
\usepackage{amssymb}
\usepackage{amsmath}
\usepackage{algorithm}
\usepackage{algorithmic}
\usepackage{amsthm}
\usepackage{multirow}
\usepackage{pgffor}
\usepackage{graphicx}
\usepackage{epstopdf}
\usepackage{tikz}
\usepackage{epstopdf}
\usepackage{mathtools}
\usepackage{enumitem}
\usepackage{subfigure}
\usepackage{tabulary}
\usepackage{color}
\usepackage{url}
\usepackage{flushend}
\usepackage{stfloats}
\usepackage{tcolorbox}
\usepackage[utf8]{inputenc}
\usepackage[english]{babel}
\usepackage{tabularx}
\usepackage{CJKutf8}
\usepackage{microtype}
\usepackage{textcomp}
\usepackage{booktabs}
\usepackage{colortbl}
\usepackage{xcolor}
\usepackage{soul}
\newcommand{\mn}[0]{$\textbf{MurKe}$}
\newcommand{\dn}[0]{$\text{HeadQA}$}

\AtBeginDocument{%
  \providecommand\BibTeX{{%
    \normalfont B\kern-0.5em{\scshape i\kern-0.25em b}\kern-0.8em\TeX}}}

\setcopyright{acmcopyright}
\copyrightyear{2020}
\acmYear{2020}
\acmDOI{10.1145/1122445.1122456}

\acmConference[Conference '20]{Conference '20}{October, 2020}{New York, USA}
\acmBooktitle{Conference '20, October, 2020}
\acmPrice{15.00}
\acmISBN{978-1-4503-XXXX-X/18/06}

\begin{document}
\title{Interpretable Multi-Step Reasoning with Knowledge Extraction on Complex Healthcare Question Answering}

\author{Ye Liu$^1$, Shaika Chowdhury$^1$, Chenwei Zhang$^{2}$, Cornelia Caragea$^1$, Philip S. Yu$^{1}$}
\affiliation{$^1$Department of Computer Science, University of Illinois at Chicago, IL, USA\\
$^2$Amazon, Seattle, WA, USA
}
\email{yliu279@uic.edu, schowd21@uic.edu, cwzhang@amazon.com, cornelia@uic.edu, psyu@uic.edu}


\begin{abstract}
Healthcare question answering assistance aims to provide customer healthcare information, which widely appears in both Web and mobile Internet. The questions usually require the assistance to have proficient healthcare background knowledge as well as the reasoning ability on the knowledge. Recently a challenge involving complex healthcare reasoning, {\dn} dataset, has been proposed, which contains multiple choice questions authorized for the public healthcare specialization exam. Unlike most other QA tasks that focus on linguistic understanding, {\dn} requires deeper reasoning involving not only knowledge extraction, but also complex reasoning with healthcare knowledge. These questions are the most challenging for current QA systems, and the current performance of the state-of-the-art method is slightly better than a random guess. In order to solve this challenging task, we present a \textbf{Mu}lti-step \textbf{r}easoning with \textbf{K}nowledge \textbf{e}xtraction framework (\textbf{\mn}). The proposed framework first extracts the healthcare knowledge as supporting documents from the large corpus. In order to find the reasoning chain and choose the correct answer, {\mn} iterates between selecting the supporting documents, reformulating the query representation using the supporting documents and getting entailment score for each choice using the entailment model. The reformulation module leverages selected documents for missing evidence, which maintains interpretability. Moreover, we are striving to make full use of off-the-shelf pretrained models. With less trainable weight, the pretrained model can easily adapt to healthcare tasks with limited training samples. From the experimental results and ablation study, our system is able to outperform several strong baselines on the {\dn} dataset.

\end{abstract}

\maketitle

\section{Introduction}
Neural network models have achieved great success on the recent progress of question answering (QA). In some of the popular datasets, such as SQuAD \cite{rajpurkar2016squad} and bAbI \cite{weston2015towards}, the machine can achieve near human-level performance. However, these datsets are easy for machine since the context contains the answer and often surface-level knowledge is sufficient to answer \cite{xiong2017dcn+}. The recently released {\dn} \cite{vilares2019head} is an ambitious test for AI systems. This dataset consists of 6,765 multiple choice questions authored for college students in the healthcare area to have the specialization license. The dataset contains question (containing context) with four or five candidate option choices from 6 categories including Medicine, Pharmacology, Psychology, Nursing, Biology and Chemistry. A small percentage ($\thicksim14\%$) of the Medicine questions refer to images, that provide additional information to answer correctly. These questions require sophisticated reasoning and language understanding abilities to be answered correctly, and even for humans (i.e., medical college students), these questions take over a period of one year or more for them to pass the exam. 



Compared to basic reading comprehension based QA setup where the answers to a question are usually found in the given small context,  the {\dn} setup needs to extract relevant knowledge according to the context and question. Another characteristic of this dataset is that unlike current datasets like TRIVIAQA-open \cite{joshi2017triviaqa}, SQUAD-open \cite{rajpurkar2016squad} and ARC \cite{clark2018think}, the gold document and relevant search document set are not provided for each question. This makes {\dn} represent a unique obstacle in the QA as the system now needs to search for relevant documents from the whole Wikipedia corpus. The performance of current state-of-the-art methods is only slightly better than random guess \cite{vilares2019head}. The performance degradation is mainly from failing to retrieve the relevant documents for the question answering model \cite{htut2018training}. 

In previous works, single-step retrieve-and-read question answering (QA) systems \cite{chen2017reading} failed to perform well on complex questions dataset \cite{yang2018hotpotqa, clark2018think} as the question does not contain sufficient retrievable clues and, thus, all the relevant context cannot be obtained in a single retrieval step \cite{qi2019answering}. The recently popular multi-hop based QA systems, like the HotpotQA \cite{yang2018hotpotqa} and WikiHop \cite{welbl2018constructing}, are designed such that they require to reason with information taken from more than one document in order to arrive at the correct answer. The reasoning chains in HotpotQA are well-designed by human; specifically, these datasets assume supporting documents are already obtained and the reasoning chain is generated by human along with the dataset. However, the {\dn} is a more natural dataset as it is collected from the healthcare specialization exam, so the reasoning chains are unknown and the model needs to extract the relevant healthcare knowledge by itself. 

The above mentioned differences lead to multiple challenges for {\dn}. First, finding the relevant supporting documents from the large corpus, like Wikipedia is a challenge, especially since standard IR approaches can be misled by distractions. Second, finding the multi-hop reasoning chain among \cite{yang2018hotpotqa} the plentiful documents is another challenge, since their reasoning path is unclear and thus would need to have a well understanding between the natural language texts. To solve these challenges, our proposed model, \textbf{Mu}ti-step \textbf{r}easoning with \textbf{K}nowledge \textbf{e}xtraction framework (\textbf{\mn}), solves this problem in two steps: extract relevant knowledge and reason with the background knowledge. The relevant knowledge extraction aims to narrow the document search space from the whole Wikipedia to the relevant document set by using a combination of token-level and semantic-level retrieval. In the reasoning step, it is possible that the answer might not be present in the initially selected documents or that the model would need to combine information across multiple documents \cite{lin2018denoising}. Thus, we extend the single-step retrieval to multi-step iterative selection, reformulation and entailment module. Given an input question, the selection module finds the most relevant document with the current question. The selected relevant document is sent to the reformulation module which aims to find the guided clue from the selected document and reformulate the current question. For this purpose, we use a reformulation module that is equipped with extractive reading-based attention to reformulate the question. The important pieces of the selected document are highlighted by what we call a reading-answer attention and integrated into a representation of the question via our reformulation module. This new question vector is then used by the selection model to re-rank the context. In this way, it allows the model to select new documents and combine evidence across multiple documents, which could provide the interpretability of the reasoning path. Moreover, the input of the reformulation and entailment module is the same and they can be processed in parallel. Therefore, our method is still efficient, even if the method needs to do iterative several steps.

The main contributions of this paper are: (a) a combination of token-level retrieval and semantic-level retrieval to settle down the search space as a small but sufficient document set, (b) an efficient and effective iterative retrieval-reformulation-entailment framework capable of complex healthcare reasoning, (c) a natural language question reformulation approach that guarantees interpretability in the multi-step evidence gathering process, and d) we illustrate the advantages of our model on the {\dn} dataset.

\section{Related Works}

\subsection{Question Answering with Knowledge Extraction}
Performing question-answering need knowledge extraction setting is far more challenging than its counterpart closed-domain setting as in the latter case the answer can be extracted from a pre-selected passage \cite{wang2018r}. For example, although the recently released AI2 Reasoning Challenge (ARC) dataset \cite{clark2018think} contains science-related questions, which also require powerful knowledge and reasoning, it has accompanying ARC corpus with relevant science sentences. As a result, in comparison to {\dn}, it is easier to find answers to ARC questions as the former requires large-scale search to find supporting documents, alongside a reading comprehension module to generate the answer. QA with knowledge extraction originally found answers in the large corpus of unstructured texts \cite{ chen2017reading}, but over the years many works have explored QA from knowledge bases (KB) such as Freebase \cite{ bollacker2008freebase} or DBPedia \cite{ auer2007dbpedia}. However, the main drawbacks of KBs are that they are incomplete as well as expensive to construct and maintain \cite{htut2018training}. This has rendered free-corpus such as Wikipedia as the preferred choice for knowledge source to provide additional evidence in answering questions, not to mention that it also provides up-to-date information \cite{chen2017reading}. Most pipelines base their information retrieval module returning the relevant documents on standard IR mechanisms (e.g., TF-IDF), which can fail to contain the correct answer in the ranked documents \cite{lee2018ranking}. 

\subsection{Multi-Step Datasets and Reasoning}
Instead of getting the answer from a single context, the questions in the multi-step datasets need to locate multiple contexts to get the answer. \cite{joshi2017triviaqa} developed TriviaQA containing question-answer pairs with several associated evidence documents, which requires inference over multiple sentences to answer correctly. Answering questions in the bAbI dataset \cite{weston2015towards} requires combining multiple disjoint evidence in the context, however, as the text is synthetic, it fails to completely resemble the complexity of passage structures in human-generated texts \cite{bauer2018commonsense}. The WikiHop dataset \cite{welbl2018constructing} requires more than one Wikipedia document to answer. More recently, the HotpotQA dataset \cite{yang2018hotpotqa} has gained traction in this direction. It contains crowdsourced questions with more diverse syntactic and semantic features \cite{jiang2019self}. A sequential approach is followed by Memory Network-based models to iteratively store the information gathered from passages in a memory cell \cite{sukhbaatar2015end, kumar2016ask, shen2017reasonet}. Works by \cite{de2018question, song2018exploring, cao2019bag} use graph convolutional network \cite{kipf2016semi} to do multi-hop reasoning. Reasoning chains fed into a BERT-based QA model is proposed by \cite{chen2019multi}. Although much progress has been made with large-scale reasoning datasets \cite{yang2015wikiqa, wang2007jeopardy}, these datasets contain gold document contexts corresponding to each question. When it comes to performing multi-step reasoning they still lag behind in performance \cite{yang2018hotpotqa}. Compare to those datasets, {\dn} is more difficult due to it does not have any relevant gold documents for each question and the reasoning chain is unknown.  
\subsection{Query Reformulation}
One direction of query reformulation works on reformulating queries by rewriting the query or only retaining their most salient terms. By selecting important terms from the retrieved document, Nogueira et al. \cite{nogueira2017task} uses reinforcement learning to reformulate the query to maximize the number of relevant documents retrieved. Beyond just selecting important terms, work by \cite{liu2019generative} refine the query in a well-formed way. In the multi-choices question answering domain, \cite{musa2018answering} use a sequence to sequence model to retain the most salient term and use an entailment model to set scores to each choice. 

Instead of reformulating the explicit query, another direction works on reformulating the latent query vector. Work by \cite{pfeiffer2018neural} showed that query refinement is effective for IR in the bio-medical domain. \cite{das2019multi,grail2020latent} turns the query reformulation to the multi-step setting, that retrieval and reader model iterative work. The reader model sets the document latent representation and uses that to reformulate query latent representation. Our work is in the latent vector reformulation direction which is more flexible and also contains the interpretation. 

\section{Task Definition}
In complex healthcare reasoning, we are given a question $\textbf{Q}$ containing m tokens $\textbf{Q} = [q_{1}, q_{2}, ...,$ $q_{M}]$ with a context description $\textbf{C} = [c_{1}, c_{2}, ...,$ $c_{L}]$ in the question where $L < M$. In some cases, an image $\textbf{I}$ is provided which contains information related to the question $\textbf{Q}$. The option set has $h$ candidate option choices $\mathcal{O} = \{\textbf{O}_{1}, \textbf{O}_{2}, ...,$ $\textbf{O}_{h}\}$, where each candidate option is a text with $R$ tokens $\textbf{O}_{i}=[o_{1}, o_{2},.., o_{R}]$. The goal is to select the correct answer $\textbf{A}$ from the candidate option set. For simplicity, we denote $\mathcal{X} = \{\textbf{Q}, \mathcal{O}, \textbf{I}\}$ as one data sample and denote $y = [y_{1}, y_2, \dots, y_h]$ as a one-hot label, where each $y_i = 1(\textbf{O}_i = \textbf{A})$ is an indicator function. In the training, N sets of $(\mathcal{X}, y)^N$ are given and the goal is to learn a model $f : \mathcal{X} \to y$. In the testing, we need to predict $y^{\rm test}$ given test samples $\mathcal{X}^{test}$.

We observe that the context itself is unable to provide enough clues to the correct answer. Hence, we seek to bring supporting knowledge for each data sample from the open knowledge base, like Wikipedia. In this work, our proposed $\textbf{\mn}$ model first extracts the supporting document as the background knowledge, and then finds the reasoning path among them. {\mn} extracts the question-related supporting documents set $\mathcal{D}_{N} = \{\textbf{D}_{1}, ..., \textbf{D}_{K}\}$ from the Wikipedia corpus $\mathcal{D}$ \footnote{https://dumps.wikimedia.org/}. Note that the extracted relevent document set $\mathcal{D}_{N}$ comes from a large corpus of documents $\mathcal{D}$, where $|\mathcal{D}| \gg |\mathcal{D}_{N}|$. Then the document set $\mathcal{D}_{N}$ is used as external background knowledge to predict the answer option $\textbf{O}_{i}\in \mathcal{O}$. By using the supporting document from the first part, we design the iterative selection, reformulation and entailment models to find the latent multi-step reasoning chain. 

\section{The Proposed Framework (MurKe)}
Since there are no correct search documents for each question in the {\dn} dataset, to get the background knowledge the model needs to search the supporting documents from the whole Wikipedia. However, this is computationally expensive and time-costing. To solve this problem, we need to settle down the space of the supporting documents such that they can cover the information of the question as much as possible, while keeping the size of the supporting document set acceptable enough for downstream processing.

After getting the supporting documents, \textbf{{\mn}} mimics how humans answer the complex question using background knowledge. Namely, based on the question and extracted supporting knowledge, humans first search one relevant background knowledge and decide whether the current background knowledge could answer the question. If it could, they get the answer. Otherwise humans change the current question according to the first and search a new background knowledge. Similarly, \textbf{{\mn}} seeks to find the latent reasoning path by selecting top-1 relevant document for the current question, reformulating the current question using the relevant document and, at the same time, performing textual entailment between the top-1 relevant document and the current question with the answer.

\subsection{Knowledge Extraction}
\begin{figure}[t]
\centering
\includegraphics[width=1\linewidth]{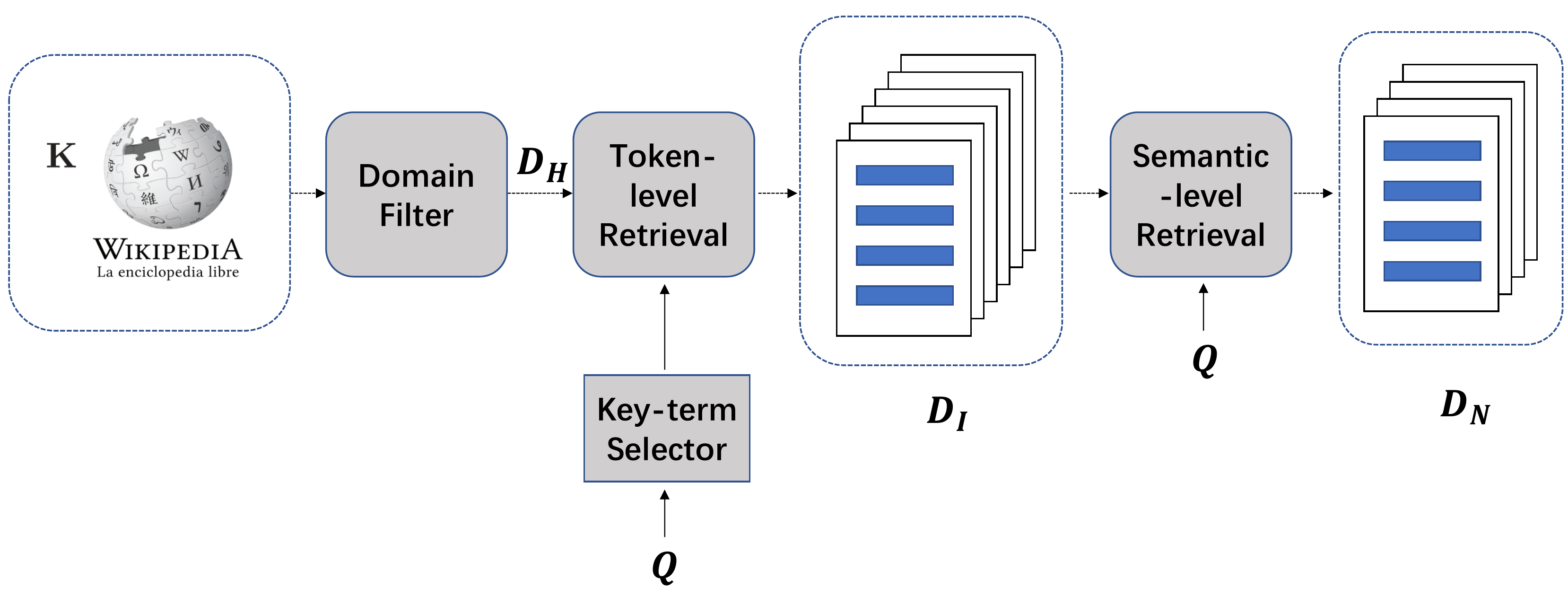}
\caption{The pipeline of the knowledge extraction step of MurKe model.}
\label{fig:proposs}
\vspace{-1em}
\end{figure}

To provide supporting documents for each healthcare question, we first introduce an effective and efficient preprocessing method to narrow the supporting document space from the whole Wikipedia to the relevant document, as  shown in Fig. \ref{fig:proposs}. Since the document scale is few hundred million, only using the semantic-level retrieval is both computation-costing and time-costing. Therefore, we use the token-level retrieval first to settle down the document number, and then use the semantic-level retrieval to further select the relevant documents. A question-based document ranking approach is employed to retrieve the most relevant supporting documents to a given question from a knowledge source Wikipedia $\mathcal{D}$. Since all the questions in our dataset are healthcare science-related, we filter the documents as the categories of depth 4 under “Health” topic \footnote{https://github.com/attardi/wikiextractor/blob/master/categories.filter}.  We refer to this corpus of extracted Wikipedia documents as the “WikiHealth” corpus $\mathcal{D}_{H}$.  

\textbf{Token-level Retrieval} To begin with, we use a combination of the neural keyword matching method and TF-IDF method to narrow down the search document scope from “WikiHealth” corpus $\mathcal{D}_{H}$ down to a set of question related documents $\mathcal{D}_{I}$. The focus of this step aims to efficiently select a candidate set that can cover the information as much as possible while keeping the size of the set acceptable enough for downstream processing. 

Specifically, following Musa et al. \cite{musa2018answering}, we treat the key-term selector as an encoder-decoder model, which inputs the question $\textbf{Q}$ and answer choices $\{\textbf{O}_{1}, \textbf{O}_{2},..., \textbf{O}_{h}\}$, and then outputs the key-terms ${\textbf{T}_{Q}} \in \textbf{Q}$ and $\{\textbf{T}_{O_{1}}$, $\textbf{T}_{O_{2}}$,..., $\textbf{T}_{O_{h}}\}$ for the question and option choices respectively. Each question forms $h$ new queries by appending each candidate answer option to the question, $\{[\textbf{T}_{Q}, \textbf{T}_{O_{1}}],$ $[\textbf{T}_{Q}, \textbf{T}_{O_{2}}],..., [\textbf{T}_{Q}, \textbf{T}_{O_{h}}]\}$. The BM25 scoring mechanism \footnote{https://radimrehurek.com/gensim/summarization/bm25.html} is then used to retrieve the $top$-100 (tested in the experiment) possibly relevant documents to each question from the WikiHealth corpus $\mathcal{D}_{H}$. The search document space of question $\textbf{Q}$ is the unit document of the retrieved documents from the $h$ queries, represented as $\mathcal{D}_{I}$. 

\textbf{Semantic-level Retrieval} After obtaining the related documents $\mathcal{D}_{I}$ using token-level retrieval, we seek to use the semantic-level retrieval to further narrow down the documents. The outputs of the neural model are treated as the relatedness score between the input question and the documents. The scores will be used to sort and limit all the upstream documents. This step, as shown in Fig. \ref{fig:proposs}, aims to screen $\mathcal{D}_{I}$ to the semantic-level related documents $\mathcal{D}_{N}$, which can be helpful for the downstream modeling.  

From these retrieved documents, to find the document that is the most semantically related to each question, we use the language model BERT \cite{devlin2018bert} which is pretrained on biomedical corpora, namely BioBERT \cite{lee2019biobert}. We use the special token to concatenate the question and document together as the input to the BERT model:

\begin{equation}
[CLS] ~ Question ~ [SEP] ~ Document ~ [SEP]
 \end{equation}
We applied an affine layer and sigmoid activation on the last layer output of the $[CLS]$ to get the scalar value. Subsequently, the documents of each question whose value is above the document relevance threshold $th_{r}$ is considered as the search documents input to our model. In our experiment, we set the threshold $th_{r}$ as 0.9. In the end, we can get the semantic-level related documents $\mathcal{D}_{N}$ for each of the search question $\textbf{Q}$.

\begin{figure*}[t]
\centering
\includegraphics[width=0.82\linewidth]{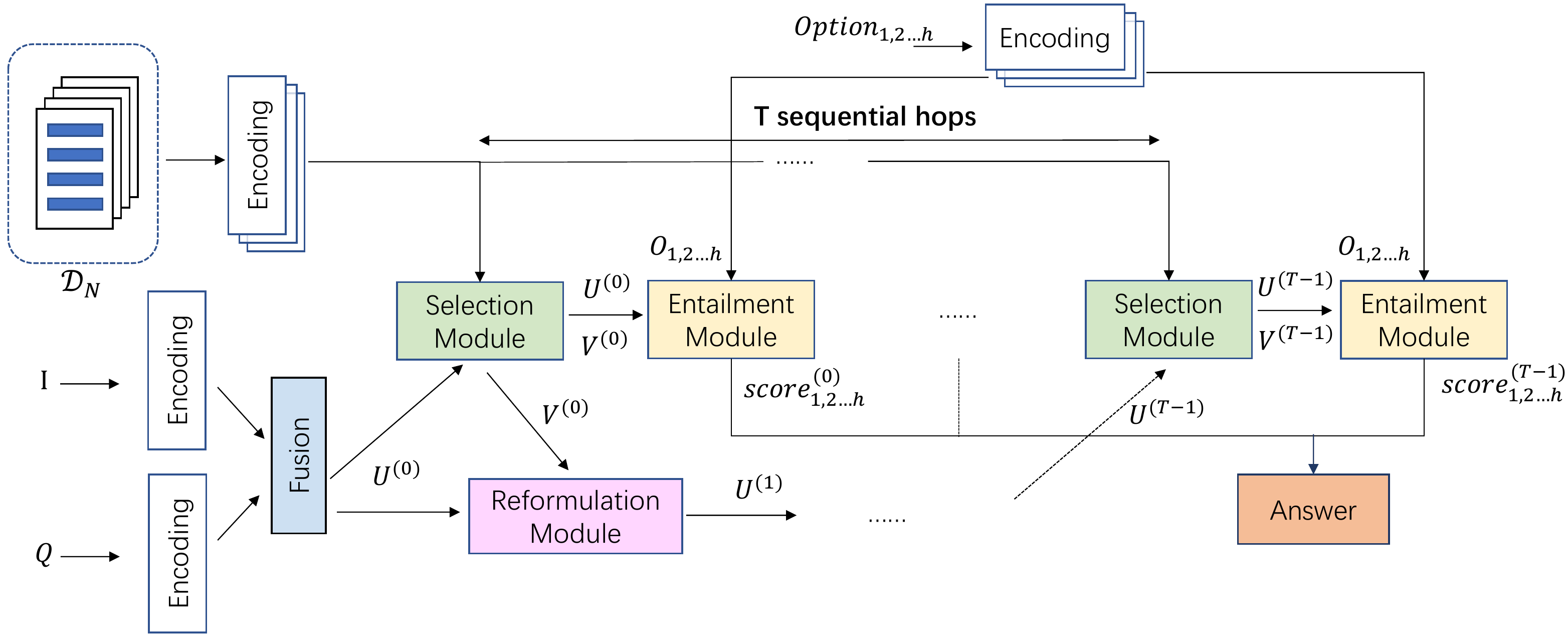}
\caption{The framework of the reasoning step of MurKe model. Selection module executes a question to obtain the most relevant documents (Sec. \ref{sec:retrieval}). The selected document is sent to both reformulation and entailment module at the same time. Reformulation module refines a question by considering the important information of selected documents (Sec. \ref{sec:reformulate}). Entailment module uses selected documents as evidence to compute probabilities for each choice (Sec. \ref{sec:entailment}) The final decision is made over time to determine the final answer (Sec. \ref{sec:final}).}
\label{fig:model1}
\vspace{-1em}
\end{figure*}

\subsection{Iterative Multi-Step Reasoning}
After getting the supporting documents $\mathcal{D}_{N}$, in this section, {\mn} proposes three modules - selection module, reformulation module and entailment module - which work iteratively to find the latent multi-step reasoning path. The reasoning diagram of the \textbf{{\mn}} framework is shown in Fig. \ref{fig:model1}. Specifically, \textbf{{\mn}} contains three sub-networks: Selection module first computes a relevance score for each document with regard to a given question and ranks them according to the score. The top one document is sent to both reformulation and entailment module. Reformulation module uses reading-answer attention to extract relevant information from top one selected document in order to update the latent representation of the question. At the same time, entailment module computes the entailment score of each candidate option and the selected document to get the final answer. Since the update of the reformulation conditions on the result of the selection module and the reformulated question can help get the confidence of the candidate choices in the following entailment model, this provides a way for the multi-step interaction between search engine (selection) and the matching model (entailment) to communicate with each other. Moreover, the model is processing very fast as the entailment model and reformulation model can be processed in parallel.
\subsubsection{Selection Module} \label{sec:retrieval}
\begin{figure*}[t]
\centering
\includegraphics[width=0.82\linewidth]{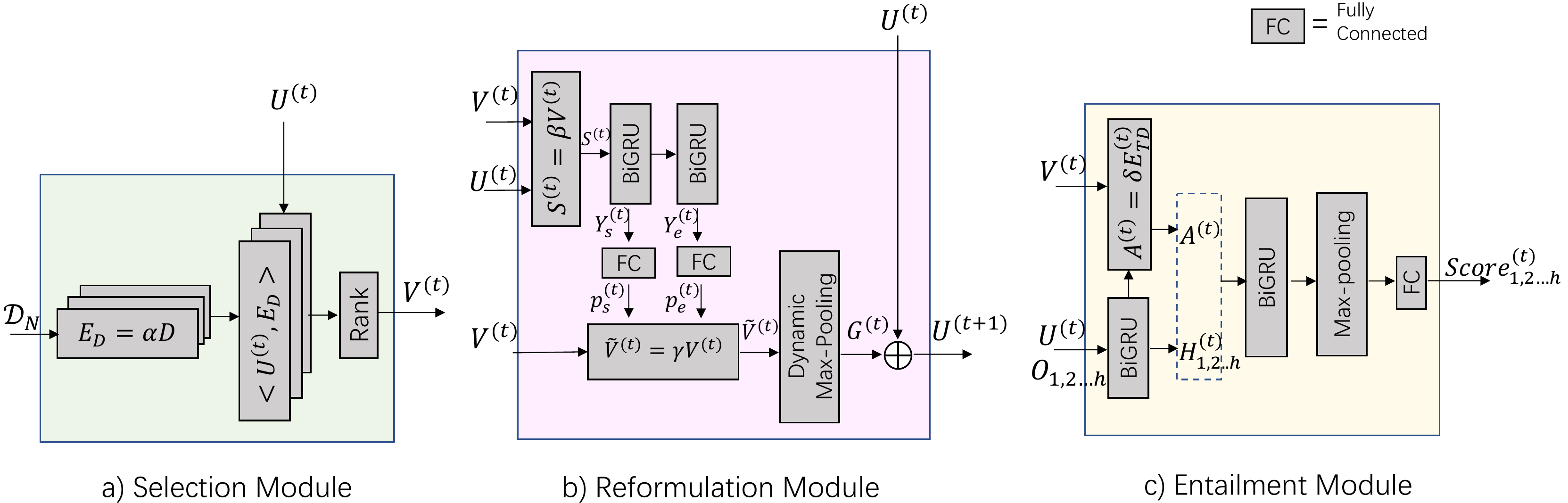}
\caption{The different building blocks of the proposed end-to-end trainable model.}
\label{fig:model2}
\vspace{-1em}
\end{figure*}

The selection module computes a relevance score between each related document and the given search question, which is represented in Fig. \ref{fig:model2} a). The related document representations are computed independently of the question and once computed, they are not updated. The relevance score of a related document is computed as an inner product between the related document and the question vectors. The related document and question representations are computed as follows. 

Given a document $\textbf{D}$ = $\rm [d_{1}, d_{2},..., d_{N}]$ in the relevant document set $\mathcal{D}_{N}$ of question $P$ consisting of $N$ tokens, a bidirectional multi-layer GRU (BiGRU) \cite{chung2014empirical} encodes each token in the document $[\textbf{d}_{1}, \textbf{d}_{2},$ $..., \textbf{d}_{N}]$ = $\rm BiGRU([d_{1}, d_{2},..., d_{N}])$, where $\textbf{d}_{j} \in \mathbb{R}^{2d}$ is the concatenation of the forward and backward GRU last layer hidden units. The question $\rm \textbf{Q} = [q_1, q_2,..., q_{M}]$ with $M$ tokens is encoded by another network with the same architecture to obtain the question embedding $\rm[ \textbf{q}_{1}, \textbf{q}_2, ..., \textbf{q}_{M}]$ = $\rm BiGRU([q_{1}, q_{2},..., q_{M}])$. To solve the long-term dependencies in the document, we compute the probability distribution $\alpha_{j}$ depending on the degree of relevance between word and the other words (in its context). The self-attention document vector $\mathbf{E_{D}} \in \mathbb{R}^{N \times 2d}$ is computed as a weighted combination of all contextual embeddings:
\begin{align}
    \alpha_{\mathbf{j}}=\frac{\exp \left(\mathbf{w} \cdot \mathbf{d}_{\mathbf{j}}\right)}{\sum_{j^{\prime}=1}^{N} \exp \left(\mathbf{w} \cdot \mathbf{d}_{\mathbf{j}^{\prime}}\right)}, 
    \quad \mathbf{E_{D}}=W_{s} \cdot \alpha_{\mathbf{j}} \cdot \mathbf{d}_{\mathbf{j}}
\end{align}
where $\textbf{w} \in \mathbb{R}^{2d}$ and $W_{s} \in \mathbb{R}^{2d \times 2d}$, used in the bilinear term, is a learned weight matrix.
In the same way, we calculate the $\rm \textbf{E}_{Q} \in \mathbb{R}^{M \times 2d} $ as the question embedding. As some queries may contain image $\textbf{I}$, so we fuse the hidden states of question embedding and the corresponding image embedding to the initial question vector as $ \rm \textbf{U}^{(0)} = Fuse(\textbf{E}_{Q}, \textbf{E}_{I})$. We tried different fusion methods, such as $\rm con()$, $\rm avg()$ and use bilinear module $\rm bil()$, which are discussed in the experiment section. 

The relevance score of a document with regard to the question ($\rm score (\mathbf{U}^{(t)},  \mathbf{E_{D}})$) is computed by a simple inner product $\langle \mathbf{U}^{(t)}, \mathbf{E_{D}} \rangle$, where $t$ means $t$-$th$ iteration. The document retriever ranks all the documents in $\mathcal{D}_{N}$ and sends the embedding of top-1 scoring document $\mathbf{E}_{D_{top1}}^{(t)}$ to the following modules, reformulation and entailment. For notation simplicity, we use $\mathbf{V}^{(t)}$ instead to refer to the top-1 scoring document $\mathbf{E}_{D_{top1}}^{(t)}$.

\subsubsection{Latent Question Reformulation Module} \label{sec:reformulate}
The latent question reformulation aims to find some evidence from the selected document so that combining the evidence with the current question representation formulates a new representation that can help answer more correctly, as shown in Fig. \ref{fig:model2} b). The reformulated question is sent back to the entailment and retriever, which uses it to calculate the entailment score between hypothesis and premise and re-rank the documents in the corpus, respectively. More formally, a reformulation module takes the encoding of the top one selected document from the previous selection module, $\mathbf{V}^{(t)}$, and the previous representation of the question, $\mathbf{U}^{(t)}$, as input, and produces an updated reformulation of the question $\mathbf{U}^{(t+1)}$. 
Moreover, in order to provide the interpretability of the model, we want to extract sub-phrase of the document which can bring the guided clue for the current question. Therefore, we formulate it as a reading comprehension task \cite{seo2016bidirectional} which aims to find the answer span in the document and use the found answer span to update the question.

\textbf{Reading-answer Attention}: The matching of stop words is presumably less important than the matching of content words. In this step, the goal is to compare the question embedding and the contextual document embeddings and select the pieces of information that are relevant to the question. We plan to learn the reading-based attention of a token as the probability that the predicted span has started before this token and will end after. Therefore, we calculate the question-aware document representation as $\rm \mathbf{S}^{(t)} = \beta_{j}\mathbf{V_{j}}^{(t)}$ where, $\rm \beta_{j} = softmax_{j} \sum_{i} \mathbf{U}^{(t)}_{i} \mathbf{W}_{c} \mathbf{V_{j}}^{(t)}$, where $\mathbf{U}^{(t)}_{i}$ represents the $i$-th token in $\mathbf{U}^{(t)}$, $\mathbf{V}^{(t)}_{j}$ represents the $j$-th token in $\mathbf{V}^{(t)}$ and the $\textbf{W}_{c}$ is the training weight. Following \cite{grail2020latent}, we use the idea of reader module to compute the reading-based attention vector. Given the question-aware document representation $\textbf{S}(t)$, we compute the starting and ending index position probability $\mathbf{p}^{(t)}_{s}$ and $\mathbf{p}^{(t)}_{e}$ using two BiGRUs followed by a linear layer and a softmax operator. They are computed from:
\begin{align} 
\mathbf{Y}^{(t)}_s &=\operatorname{BiGRU}\left(\mathbf{S}^{(t)}\right) & \mathbf{Y}^{(t)}_e &=\operatorname{BiGRU}\left(\mathbf{Y}^{(t)}_{s}\right) \\ \mathbf{p}^{(t)}_s &=\operatorname{softmax}\left(\mathbf{w}_{s} \mathbf{Y}^{(t) }_s\right) & \mathbf{p}^{(t)}_e &=\operatorname{softmax}\left(\mathbf{w}_{e} \mathbf{Y}^{(t)}_e\right) 
\end{align}

where $w_e$ and $w_s$ are trainable vectors of $\mathbb{R}^{2d}$. The two probability vectors $\mathbf{p}^{(t)}_s \in \mathbb{R^{N}}$ and $\mathbf{p}^{(t)}_e \in \mathbb{R^{N}}$ are not used to predict an answer, but to compute a reading-based attention vector $\gamma^{(t)}$ over the document. Intuitively, these probabilities represent at step $t$ how likely each word is to be the beginning and the end of the answer span respectively. We define the reading-based attention of a token as the probability that the predicted span has started before this token and will end after, which can be computed as follows:
\begin{align}
     \gamma_{i}^{(t)}=\left(\sum_{k=0}^{i} {\mathbf{p}_{s_{k}}^{(t)}}\right)\left(\sum_{k=i}^{N} {\mathbf{p}_{e_{k}}^{(t)}}\right)
\end{align}
Further, we use these attention values to re-weight each token of the document representation. We compute $\widetilde{\mathbf{V}_{i}}^{(t)} \in \mathbb{R}^{N\times2d}$ with:
\begin{align}
    {\widetilde{\mathbf{V}_{i}}^{(t)}} = \gamma_{i}^{(t)} \mathbf{V_{i}}^{(t)}
\end{align}
\textbf{Dynamic Knowledge Extraction Max-Pooling}: This layer aims at collecting the relevant evidences of $\widetilde{\mathbf{V}_{i}}^{(t)}$ with length $N$ to add to the current question representation with length $M$. We partition the row of the initial sequence into $M$ approximately equal parts. It produces a grid of $M \times 2d$ in which we apply a max-pooling operator in each window. As a result, a matrix of fixed dimension adequately represents the input, preserving the global structure of the document, and focusing on the important elements of each region. This can be seen as an adaptation of the dynamic pooling layer proposed by Socher et al. \cite{socher2011dynamic}. Formally, let $\widetilde{\mathbf{V}_{i}}^{(t)}$ be the input matrix representation, we dynamically compute the kernal size, $w$, of the max-pooling according to the length of the input sequence and the required output shape: $w = \lceil \frac{N}
{M} \rceil$, $\lceil \cdot \rceil$ being the ceiling function. Then the output representation of this pooling layer is the extracted knowledge from the document represented as $\textbf{G}^{(t)} \in \mathbb{R}^{M\times2d}$ where
\begin{align}
    \textbf{G}_{i}^{(t)}=\max_{k \in\{i w, \ldots,(i+1) w\}}\left(\widetilde{\mathbf{V}_{k}}^{(t)}\right)
\end{align}
Finally, the updated representation of the question $\mathbf{U}^{(t+1)} \in R^{M\times2d}$ is the sum of $\mathbf{U}^{(t)}$ and $\mathbf{G}^{(t)}$. 

\subsubsection{Entailment Module} \label{sec:entailment}
Given the selected top-1 document, the module needs to select a particular answer from the candidate options choices $\mathcal{O}=\{\textbf{O}_{1}, \textbf{O}_2, ...,$ $\textbf{O}_{h}\}$. In the approach pioneered by \cite{khot2018scitail}, a multiple reading comprehension problem is converted into an entailment problem wherein each top-1 selected document is a premise and the question combined with each candidate answer is used as a hypothesis, and the model’s probability that the premise entails this hypothesis becomes the candidate answer’s score. 

In our method, the embedding of selected top-1 document $\mathbf{V}^{(t)}$ is treated as the embedding vector of the premise $\textbf{P}^{(t)}$, while the hypothesis is the combination of question and candidate choices. Since the question representation is vectorized as $\textbf{U}^{(t)}$ but the candidate choices is still in the token-level, 
the question embedding $\textbf{U}^{(t)}$ is treated as the initial hidden states and the choice token $\mathcal{O}=\{\textbf{O}_{1}, \textbf{O}_{2}, ... \textbf{O}_{h}\}$ are passed through a BiGRU separately in order to capture the dependency between the words, which returns a new hypothesis representation $\textbf{H}^{(t)}_{1,2...,h} \in  \mathbb{R}^{2d\times(M+len(\textbf{O}_{1,2...,h}))}$. Then, an attention mechanism is used to determine the attention weighted representation of the $j$-th word in the premise as follows:
\begin{align}
    e_{ij} = \textbf{P}^{(t)}_{i} \cdot \textbf{H}^{(t)}_{j}, ~\delta_{ij}=\frac{\rm exp(e_{ij})}{\sum_{r=1}^{K} \rm exp(e_{rj})},~ \textbf{A}^{(t)}_{j} = \sum_{i} \delta_{ij} \textbf{P}^{(t)}_{i}
\end{align}
The matching layer is a $\rm BiGRU(\cdot)$ with the input $\textbf{M}^{(t)}_{j} = [\textbf{A}^{(t)}_{j}; \textbf{H}^{(t)}_{j}]$ ($[;]$ is the concatenation operator). Finally, the max-pooling result over the hidden states of the matching model is used for softmax classification to get the entailment score for each choice $\rm \{Score_{1}^{(t)}, Score_{2}^{(t)}$ $, \cdots,\rm Score_{h}^{(t)}\}$. The diagram of entailment module is Fig. \ref{fig:model2} c). It is worth to notice that the input of the entailment module and question reformulation module are the selected top-1 document $\textbf{V}^{(t)}$ and the latent question representation $\textbf{U}^{(t)}$, so those two models can be processed in parallel.

\begin{figure*}[t]
\centering
\includegraphics[width=0.85\linewidth]{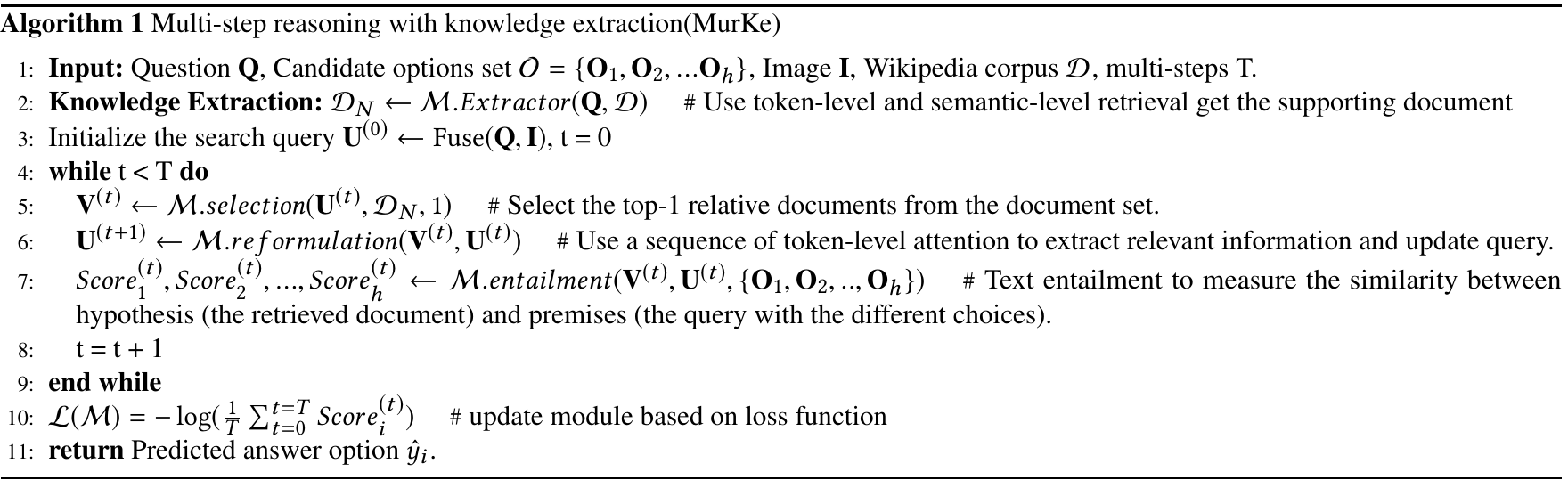}
\vspace{-1em}
\label{fig:algo1}
\end{figure*}

\subsubsection{Multi-Step Iterative Reasoning with knowledge extraction} \label{sec:final}
Our multi-step reasoning architecture with knowledge extraction is summarized above in Algorithm 1. Given a large-scale text corpus (as Wikipedia), the question text and candidate option set, our model returns the predicted answer choice. To narrow down the search document from the whole Wikipedia to the question relevant document set, we use the token-level retrieval followed by the semantic-level retrieval (line 3). The multi-step interaction between the selection, reformulation and entailment model can be best understood by the while loop from line 4 to line 9. The initial question $\textbf{U}_{0}$ is first used to rank all the documents in the relevant documents (line 5), followed by which the $top$-1 document is sent to both the reformulation model and the entailment model. The reformulation uses the selected document to calculate the important span in the document and update the question representation $\textbf{U}_{t+1}$ (line 6). The entailment model treats the $top$-1 selected document as the premise and the combination of question and candidate choices as the hypothesis to compute the entailment score for each choice (line 7). The updated question is sent back to the selection module (line 5). The selection module uses this updated question to re-rank the documents and the entire process is repeated for T steps. 
At the end of T steps, the model returns the choice with the score returned by T steps ${Score_{1}^{(t)}, Score_{2}^{(t)}, ..., Score_{h}^{(t)}}$. We update the model using log likelihood as the objective function (line 11):
\begin{align}
    \mathcal{L} = - \log(\frac{1}{T}\sum_{t=0}^{t=T}Score_{i}^{(t)})
\end{align}
During inference, based on the multi-step score of each choice aggregation, the choice with the highest score is the predicted answer. 

\section{Experiment}
We now present experiments to show the effectiveness of each component of our framework. We test the {\dn} dataset both on supervised and unsupervised settings, and further perform ablation study to evaluate the contribution from each part of the model. In the end, we present a case study to demonstrate the interpretability of our model. 

\subsection{Dataset}

\begin{table}[]
\centering
\resizebox{2.5in}{!}{
\begin{tabular}{l|ccc|c}
\toprule
                    &  \multicolumn{3}{l|}{Supervised setting} & \multicolumn{1}{l}{Unsupervised} \\
Category            & Train       & Dev         & Test &     Setting           \\ \midrule
Biology (BIR)   &452 &226 &454 &1,132     \\
Nursing (EIR)     &384 &230 &455 &1,069   \\
Pharmacology (FIR) &457 &225 &457 &1,139   \\
Medicine (MIR)  &455 &231 &463 &1149   \\
Psychology (PIR)  &453 &226 &455 &1134    \\
Chemistry (QIR)  &456 &228 &458 &1142 \\
\hline
Total  &2657 &1366 &2742 &6765   \\
\bottomrule
\end{tabular}
}
\caption{Data Statistics Summary of {\dn}}
\label{tab:Table1}
\vspace{-2em}
\end{table}

\textbf{{\dn}} This dataset is created from examinations, spanning the years 2013 to 2017, that are designed for obtaining specialization positions in the Spanish public healthcare areas. It contains graduate-level multi-choice questions about Medicine (MIR), Pharmacology (FIR), Psychology (PIR), Nursing (EIR), Biology (BIR), and Chemistry (QIR). The original version of this dataset is in Spanish, but it has also been translated to English. We use the English version of this dataset. There is a total number of 6765 question-answer pairs and the questions in the Medicine category (MIR) ($\thicksim14\%$ among all questions) have image, which we use in question initialization. Table \ref{tab:Table1} summarizes the number of questions in each category and the data splits. 

The dataset has supervised and unsupervised settings. In the supervised setting, exams from 2013 and 2014 are used for the training set, 2015 for the development set, and the rest for testing. In the unsupervised setting, we pre-trained the model on other similar tasks or datasets and test the performance of the whole dataset.

\begin{table*}[]
\resizebox{0.69\textwidth}{!}{
\begin{tabular}{cccccccc|ccccccc}
\toprule
\multicolumn{1}{l|}{}                     & \multicolumn{1}{l}{} & \multicolumn{1}{l}{} & \multicolumn{1}{l}{} & \multicolumn{1}{l}{\textbf{Acc}} & \multicolumn{1}{l}{} & \multicolumn{1}{l}{} & \multicolumn{1}{l|}{} &     &     &     & \textbf{Point} &     &     &     \\ \hline
\multicolumn{1}{c|}{Models}              & BIR                  & MIR                  & EIR                  & FIR                     & PIR                  & QIR                  & Avg                  & BIR & MIR & EIR & FIR   & PIR & QIR & Avg \\ \hline
\multicolumn{1}{c|}{DRQA}           & 29.5        & 25.0      & 27.3      & 28.3      & 31.0     & 30.2         & 28.5                
& 40.8 & -0.2 & 20.6 & 29.8 & 54.0 & 47.6 & 32.1    \\
\multicolumn{1}{c|}{BIDAF}       & 33.4       & 26.2       & 26.8       & 29.9         & 26.8      & 30.3      & 28.9                 & 75.6 & 11.0 & 15.8 & 44.4 & 16.6 & 48.6 & 35.3     \\
\multicolumn{1}{c|}{DGEM}         & 31.7      & 25.7        & 28.7      & 29.8     & 28.5       & 30.3      & 29.1                
 & 60.8 & 7.0 & 34.2 & 45.0 & 31.6 & 48.4 & 37.8    \\
\multicolumn{1}{c|}{DECOMPATT}    & 30.6      & 23.6        & 27.9      & 27.2      & 28.3      & 27.6    & 27.5                
& 51.2 & -13.0 & 27.8 & 20.2 & 30.0 & 23.6 & 23.3     \\
\multicolumn{1}{c|}{TFIDF-IR}     & 37.9       & 30.3       & 32.6       & 38.7     & 34.7        & 33.7       & 34.6                 & 116.8 & 48.6 & 67.8 & 125.0 & 87.6 & 79.6 & 87.6      \\
\multicolumn{1}{c|}{IR + BERT}   & 29.6          & 31.0      & 33.8       & 33.7       & 30.0      & 33.9      & 32.0                 & 41.6 & 55.0	& 76.6	& 79.4	&45.2	&82.0 & 63.3    \\
\multicolumn{1}{c|}{IR + BioBERT}   & 34.3       & 32.3       & 32.5       & 31.7        & 32.8     & 31.1      & 32.4                &  84.0 & 67.0	& 67.0	& 61.0	& 70.8	&55.6	& 67.6      \\
\multicolumn{1}{c|}{Multi-step TFIDF-IR} &  35.6 &   32.7  & 33.5       & 35.3        & 36.4     & 33.3      & 34.9                & 102.2 &  74.4  &  78.2 & 100.2 &  107.6 &  79.6 &  89.3    \\
\multicolumn{1}{c|}{Multi-step Reasoner} & 39.7   & 40.1     & 40.2     & 41.3     & 44.0      & 43.0      & 41.7                
& 135.0  & 132.6 & 137.4 & 138.6 & 175.6 &  164.4  & 151.3     \\ \hline
\multicolumn{1}{c|}{\textbf{{\mn}}}       & \textbf{45.5}  & \textbf{42.4}         & \textbf{42.3}  & \textbf{48.0}        & \textbf{44.3}       & \textbf{44.3}    & \textbf{44.4}                 
& \textbf{189.4} & \textbf{158.8}  & \textbf{158.8} & \textbf{209.6} & \textbf{160.6} & \textbf{173.0}   & \textbf{172.3}    \\
\bottomrule
\end{tabular}
}
\caption{Accuracy and POINTS on the {\dn} corpora (unsupervised setting)}
\label{unsup}
\vspace{-1em}
\end{table*}

\begin{table*}[]
\resizebox{0.69\textwidth}{!}{
\begin{tabular}{cccccccc|ccccccc}
\toprule
\multicolumn{1}{l|}{}                     & \multicolumn{1}{l}{} & \multicolumn{1}{l}{} & \multicolumn{1}{l}{} & \multicolumn{1}{l}{\textbf{Acc}} & \multicolumn{1}{l}{} & \multicolumn{1}{l}{} & \multicolumn{1}{l|}{} &     &     &     & \textbf{Point} &     &     &     \\ \hline
\multicolumn{1}{c|}{Models}              & BIR                  & MIR                  & EIR                  & FIR                     & PIR                  & QIR                  & Avg                  & BIR & MIR & EIR & FIR   & PIR & QIR & Avg \\ \hline
\multicolumn{1}{c|}{BIDAF}       & 36.5 & 26.6 & 27.7 & 29.3 & 28.1 & 34.1 & 30.3                                              & 104.0 & 14.5& 18.5& 39.0& 29.0& 83.0 & 48.0    \\
\multicolumn{1}{c|}{DGEM}        & 31.7 & 27.2 & 30.7 & 29.9 & 31.0 & 33.2 & 30.6              
& 61.0 & 20.5 & 52.5 & 45.5 & 54.5& 75.0 & 51.5    \\
\multicolumn{1}{c|}{TFIDF-IR}     & 39.8 & 33.3 & 36.4 & 42.2 & 35.7 & 36.0 & 37.2                                             & 116.8 & 48.6 & 67.8 & 125.0 & 87.6 & 79.6 & 87.6      \\
\multicolumn{1}{c|}{IR + BERT}   & 35.2	& 35.6	& 38.2	& 33.7	& 33.7	& 33.4 	& 35.0                                         & 92.8 & 97.4	& 113.4	& 79.4	& 78.8	& 77.2	& 89.8    \\
\multicolumn{1}{c|}{IR + BioBERT}   & 38.0	& 36.4	& 37.8	& 33.7	& 33.6	& 38.9	& 36.4                                     & 104.6 & 111.0	& 48.6	& 79.4	& 78.0	& 127.6	& 103.0      \\
\multicolumn{1}{c|}{Multi-step TFIDF-IR} & 41.9 & 38.1 & 36.6 & 39.2 & 40.3 & 39.1 & 39.2                                 
& 155.0 & 118.4& 99.0  &129.8 & 138.8 &129.2 & 128.4     \\
\multicolumn{1}{c|}{Multi-step Reasoner} & 43.4 & 42.9 & 42.9 & 43.7 & 43.5 & 44.3 & 42.9                                  & 162.4 & 178.2 & 159.0 & 170.6 & 162.0 & 163.6 & 166.0     \\ \hline
\multicolumn{1}{c|}{\textbf{{\mn}}}       & \textbf{47.1}   & \textbf{45.6}        & \textbf{46.7}       & \textbf{48.8}     & \textbf{46.7}         & \textbf{45.5}         & \textbf{46.7}                 
& \textbf{200.0}  & \textbf{189.4} & \textbf{184.6} & \textbf{217.0} & \textbf{197.2}  & \textbf{186.8}  & \textbf{199.8}    \\
\bottomrule
\end{tabular}
}
\caption{Accuracy and POINTS on the {\dn} corpora (supervised setting)}
\label{sup}
\vspace{-1em}
\end{table*}
\subsection{Evaluation on Reasoning Ability}
In this section, we evaluate the performance of reasoning ability of {\mn} on the {\dn} data. Since the {\dn} data is very small, we use the other related datasets to pre-train the different modules of {\mn}. Recent studies \cite{howard2018universal, devlin2018bert} have shown the benefit of fine-tuning on similar tasks or datasets for knowledge transfer. Considering the unique challenge of HeadQA, we explore the related retrieval, reading comprehension and entailment task-specific datasets for knowledge transfer. We directly adapt the pre-train weight without further training on the {\dn} dataset, which is defined as the unsupervised setting. In the supervised setting, we  initialize {\mn} with the pre-trained weight and then finetune on {\dn}.

\textbf{Metrics} We use Accuracy (Acc) and POINTS metric (used in the official exams): a right answer counts 3 points and a wrong one subtracts 1 point. \footnote{Note that as some exams have more choices than others,
there is not a direct correspondence between accuracy and
POINTS (a given healthcare area might have better accuracy
than another one, but worse POINTS score).}

\subsubsection{Training Details}
All the bi-directional GRU are with a single hidden layer (d = 200). The input of BiGRU at each token is the pre-trained BioWordVec embedding (200-dimensional) \footnote{https://github.com/ncbi-nlp/BioSentVec}, which trains the word embedding using PubMed \footnote{https://www.ncbi.nlm.nih.gov/pubmed/} and the clinical notes, and this BioWordVec covers $98\%$ words in our dataset. Additionally, to capture the structural representation of the words, we incorporate the background knowledge in the form of graph embedding using the ConceptNet \cite{speer2017conceptnet} knowledge base. In the end, both embeddings are concatenated to form the final word embeddings (300-dimensional). 

In the unsupervised setting, the BiGRU in the retrieval model is pre-trained with the document and question encoder-decoder model. The reformulator is pre-trained using supervised learning (using the correct spans as supervision), where we use SQuAD data \cite{rajpurkar2016squad} to train this model as the pre-trained weight. The entailment model is trained using entailment task-specific dataset SciTail \cite{khot2018scitail}. In the supervised setting, we first train the model by setting the number of multi-step iterative method steps (T = 1), then we train the model with different step numbers. We train the models for 50 iterations using SGD with a learning rate of 0.015 and learning rate decay of 0.05.

\subsubsection{Baselines}
\textbf{DrQA} \cite{ chen2017reading} consists of a Document Retriever module based on bigram hashing and TF-IDF matching to return the five most relevant Wikipedia articles to a given question and a machine comprehension module, that is implemented using a multi-layer recurrent neural network trained on SQuAD \cite{rajpurkar2016squad} to find the exact span containing the correct answer. Similar to \cite{vilares2019head}, the answer containing the most overlapping tokens with the selected span is considered as the correct answer for the multi-choice setting. To apply to our dataset, we calculate the similarity between selected spans and options. The option with the highest score is treated as the correct answer. \textbf{BiDAF}  \cite{seo2016bidirectional} the document retriever is the same as DrQA \cite{chen2017reading}, but the Document Reader uses bi-directional attention flow mechanism and hierarchical embedding process to obtain a question-aware context representation that is used to predict the correct answer span.  
\textbf{DecompAttn} \cite{khot2018scitail} is a textual entailment system that first forms hypothesis by appending each candidate answer option to the question. The hypothesis is then used in turn as a question to retrieve relevant sentences to be considered as the premises. The degree of a premise entailing a hypothesis is then computed as the entailment score and the answer in the hypothesis leading to the highest score is the correct answer. \textbf{DGEM} \cite{parikh2016decomposable} is a neural attention-based entailment system that decomposes the task into sub-problems to be solved in a parallelizable manner, where the results are merged to produce the final classification output. 
\textbf{TFIDF Retrieval}, which is similar to the IR baselines by \cite{clark2016combining, clark2018think}, uses the DrQA \cite{chen2017reading} 's Document Retriever, which scores the relation between the queries and the articles as TF-IDF weighted bag-of-word vectors, and also takes into account word order and bi-gram counting. The predicted answer is defined as the one having the maximum score in the question for which we obtained the highest document relevance. \textbf{Retrieval + BioBERT$\|$BERT entailment model} uses the top one document from the retrieval module and pre-trained BioBERT$\|$BERT to get the entailment score between premise (search document) and hypothesis (question with the choice), where the top one hypothesis is the answer. \textbf{Multi-step TF-IDF retrieval (using keywords)} uses the keywords obtained from the previous document to reformulate the question and then uses TF-IDF to retrieve the new document. \textbf{Multi-step-reasoner} \cite{das2019multi} the multi-step framework using reinforcement learning (RL) where retriever and reader (DrQA) iteratively interact with each other to get the final answer. 

\subsubsection{Results}
\quad \textbf{Unsupervised Setting} Table \ref{unsup} shows the accuracy and POINTS scores for {\dn}. Our model {\mn} performs best among the baselines. As can be seen, even a powerful model like BERT performs unsatisfactorily on the {\dn} dataset. The main reason is that the initial question does not contain sufficient retrievable clues to find the document containing the answer. Whereas, the multiple steps of iterative reasoning of our proposed model help to reformulate the question with the missing information, which in turn facilitates in retrieving the document related to the answer and uniformly increases performance over the base model. Moreover, using different task-related datasets to pre-train each module separately is promising to achieve an acceptable performance. \\
\textbf{Supervised Setting} We show the performance of the top models on the test split corresponding to the supervised setting in Table \ref{sup}. Our proposed model {\mn} performs substantially better than the other baselines, which shows that by using the multi-step iteration it is possible to have the model better match the gold document and get a better entailment score. The other multi-step methods, like multi-step TFIDF-IR and Multi-step Reasoner perform worse than our method. This is primarily because the multi-step TFIDF-IR methods rely on statistical features like frequency of terms in the document, and fail to explicitly use information about entities that may not be frequently occurring in the document. We also find that RL approaches, Multi-step Reasoner, are slow to converge as rewards from a down-stream task are sparse and action space in information retrieval is very large. Table \ref{human} shows humans performance. The first row is the average of the top 10 scores gotten by humans and the second row is the passing score, meaning that the examinee can pass the exam if it receives above this score. Compared to the score of the pass mark, {\mn} passes three categories (EIR, PIR, and QIR) and the avg point is higher than the pass mark. Nevertheless, there is still a long way to beat the best performance of humans. 

\begin{table}[]
\resizebox{0.43\textwidth}{!}{
\begin{tabular}{c|ccccccc}
\toprule
          & BIR  & MIR  & EIR  & FIR  & PIR  & QIR & Avg  \\
\hline
Avg 10 best humans     & 627.1 & 592.2 & 515.2 & 575.5 & 602.1 & 529.1 & 477.6\\
Pass mark      & 219.0 & 207.0 & 180.0 & 201.0 & 210.0 &185.0 & 200.3\\ \hline
\textbf{{\mn}}  & 199.8 & 196.2 & 215.4 & 196.4 & 217.2 & 203.7& 204.8\\
\bottomrule
\end{tabular}
}
\caption{Human performance on the 2016 exams (Points).}
\label{human}
\vspace{-2em}
\end{table}
\vspace{-2em}
\subsubsection{Influence of the number of Reasoning Step}
\begin{figure}[t]
\centering
\includegraphics[width=0.6\linewidth]{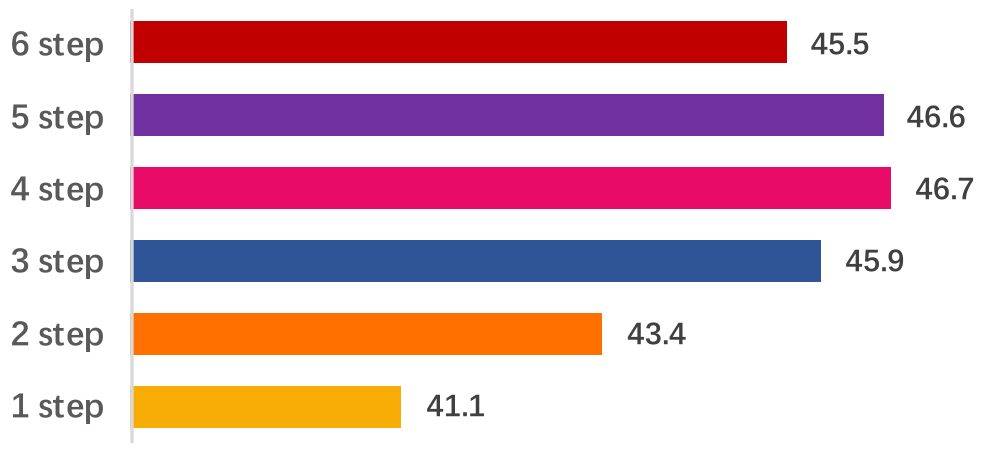}
\caption{The accuracy with the different reasoning step.}
\label{figure:step}
\vspace{-1em}
\end{figure}

As we can see from Fig. \ref{figure:step}, without the multi-step (using 1 step), the performance is very poor $41.1$. By increasing steps of interaction, the quality in terms of answer accuracy becomes better, which indicates that even though the correct document (containing the answer string) was not retrieved in the first step, the retriever is able to fetch relevant documents later. The performance keeps increasing with the increase of the iterative step. But when the number of steps is too high (6 steps), the performance declines, which may indicate that more noises are added with more steps. Therefore, in most cases the optimal value of $T$ lies in a small range of values as demonstrated in \cite{das2019multi} and it is not time-consuming to find it using the grid search strategy in practical applications. It is also unsurprising to see that when correct documents are retrieved, the performance of the entailment model also increases and it is easy to find the correct final answer.



\begin{table}[]
\resizebox{0.32\textwidth}{!}{
\begin{tabular}{ccccc}
\toprule
             &           & F1-score & Accuracy & Points \\ \hline
Unsupervised   & w/o       &  42.2    &     42.5     &    158.5  \\
              & avg       &   42.5       &  42.7    &  163.4    \\  \hline
Supervised   & w/o       &   44.8       &    45.0      &   181.4    \\
             & bil &     45.7     &     45.9     &  188.8   \\ 
             &  con       &    45.3      &    45.5      & 186.7   \\
\bottomrule
\end{tabular}
}
\caption{Different fusion methods on both unsupervised and supervised settings on MID data. Here avg stands for average and con for concatenate.}
\label{figure:fusion}
\vspace{-1em}
\end{table}

\subsubsection{Performance using multi-modality fusion}
We also test the performance of the proposed model using different multi-modality fusion methods on the questions in the Medicine category (MIR) which have image. Among all the questions, 101 questions have image information. For the image embedding, we first loaded the pre-trained Resnet \cite{he2016deep} model with 18 layers \footnote{https://github.com/qubvel/classification$\_$models}, removed its final output layer (the softmax layer), added new layers ( flatten layer, dense fully connected layer of 200 units and output layer to predict probs of 10 classes). We train the weights of the pre-trained model with new layers together and then remove the output layer and extract features from the dense layer of 200 units as the embedding of each image. In the unsupervised setting, we compare without fusion method (w/o) and average the question embedding and image embedding (avg). In the supervised setting, we evaluate without fusion method (w/o), use the bi-linear model (bil) and concatenate the question embedding and image embedding and project to the pre dimension (con).

As seen from Table \ref{figure:fusion}, fusing additional image information with the question representation can help improve the performance. It is not surprising that the question is related to the image, so by adding image information, it helps the model to better retrieve the document. In the supervised setting, the pre-trained weight can help the image embedding learn a projection from the image space to the text space, so it has more improvement.

\subsection{Evaluation on Knowledge Extraction}
In this section, we want to demonstrate the performance of the knowledge extraction which combines token-level retrieval and semantic retrieval, and indicates the necessity of using multi-step reasoning. We use the NCRF++ \cite{yang2018ncrf++} to get the selected question key-terms and is trained on the Essential Terms dataset \footnote{https://github.com/allenai/essential-terms} introduced by Khashabi et al. \cite{khashabi2017learning}. Similarly, we use the same processing on the answer. We calculate the amount that both the question and answer appear in the same document. We find that only 21 questions (6765 in total) have one document that contains both question and answer key term, which shows the complexity of the question and the importance of using multi-step iterative method to solve this problem. 


\begin{figure*}[t]
\centering
\includegraphics[width=0.88\linewidth]{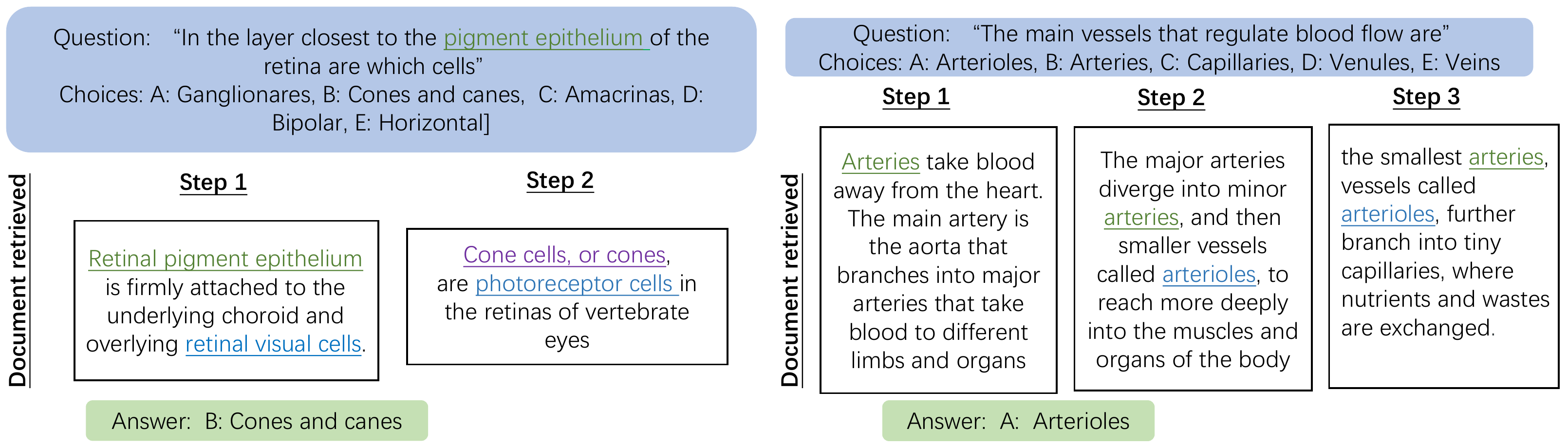}
\caption{Examples of how multi-step reasoner iteratively modifies the question by reading context to find more relevant documents.}
\label{figure:case}
\vspace{-1em}
\end{figure*}

\newcommand{\hlcr}[2][yellow]{{%
    \colorlet{foo}{blue!#1}%
    \sethlcolor{foo}\hl{#2}}}

\begin{figure}[t]
\centering
\includegraphics[width=0.65\linewidth]{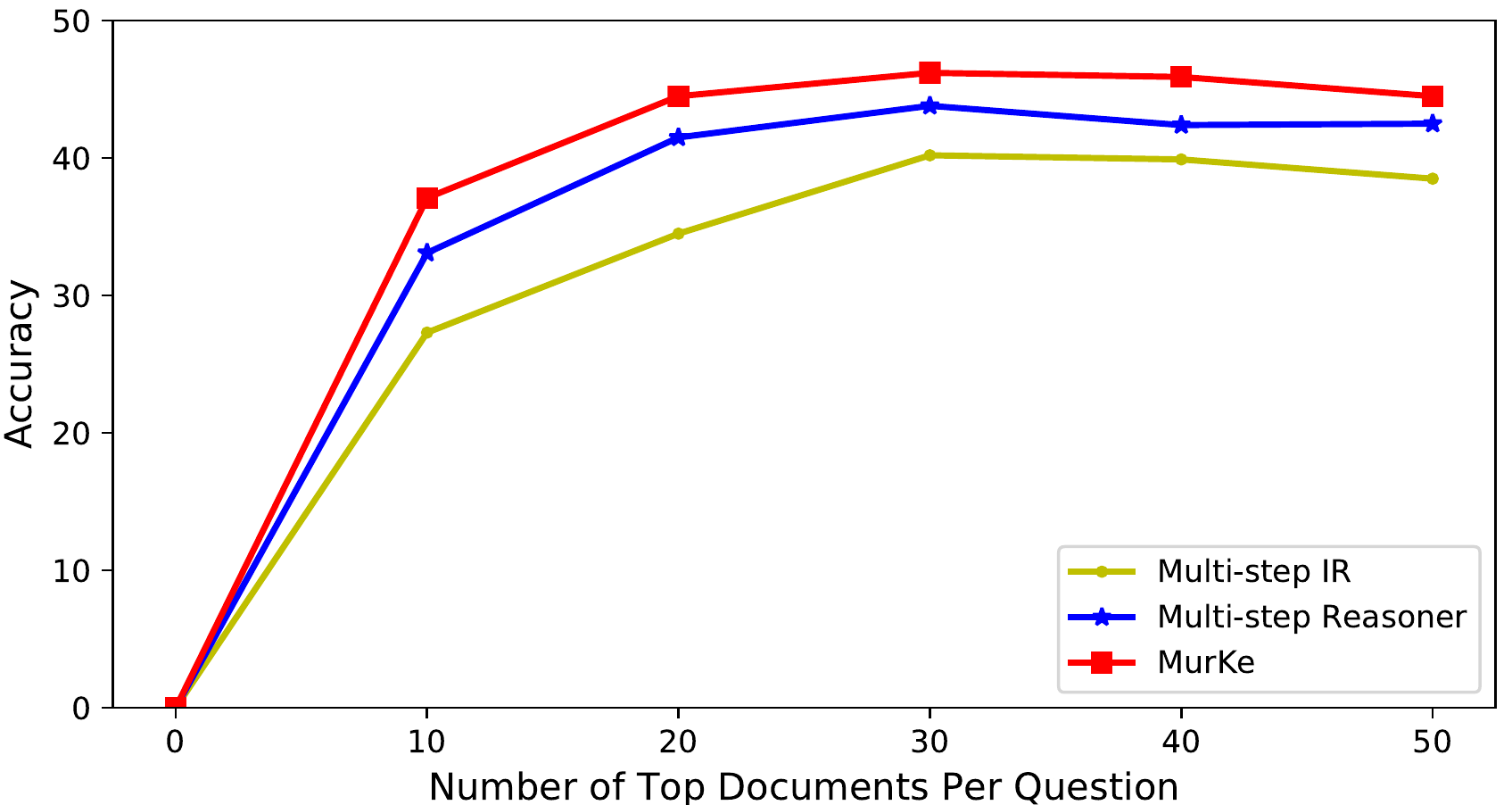}
\caption{The accuracy with the different search document number.}
\label{figure:search_num}
\vspace{-1em}
\end{figure}

\subsubsection{Influence of different supporting document scale}
We want to narrow the supporting document space that not only contains the information needed by question and answer, but also will not be redundant. To know the influence of supporting document scale, instead of using threshold after the semantic-level retrieval, we rank the supporting documents and select the scale from the top 10 to top 50 to assess the performance. From Fig. \ref{figure:search_num}, we can see that with the growth of supporting documents of question, the performance gets better. When the number of documents is around top 30, all methods reach the best performance. However, when the number of supporting documents goes beyond that, the performance shows a little bit drop. This may be because with the increase of documents, it leads to more deceptive documents. 

\subsection{Interpretable Ability of MurKe}
Fig. \ref{figure:case} shows two instances where iterative interaction is helpful. In the left figure, the retriever is initially unable to find a document that can directly answer the question. However, it finds a document which has a different description of ``visual cells'', ``photoreceptor cells'', allowing it to find a more relevant document that directly answers the question. In the figure at the right side, the retrieved documents indicate that both ``Arterioles'' and ``Arteries'' could be the answer. Based on the fact that the smallest ``arteries'' is ``arterioles'', which reach into the muscles and organs of the body, so ``arterioles'' is the main vessels that regulate the blood flow. Since we aggregate (sum) the scores of entailment of each retrieved documents with choices, this leads to an increase in the score of the choice (``Arterioles'') to be the predicted answer. Therefore, by using reading-answer attention in the reformulation module, our module {\mn} can clearly show us which part in the document is highlighted as the clue regarding current question and provides the interpretability of the reasoning path. 




\balance
\section{Conclusions}
In this paper, we present a system {\mn} that answers healthcare exam questions by using knowledge extraction and multi-step reasoning. To get a relevant document for each question, {\mn} retrieves supporting documents from a large, noisy corpus on the basis of keywords extracted from the original question and semantic retrieval. {\mn} proposes the multi-step iterative method to solve complex healthcare QA, which uses information selected by combining iterative question reformulation and textual entailment. Our neural architecture uses a sequence of token-level attention mechanisms to extract relevant evidence from the selected documents in order to update the latent representation of the question, which shows the interpretability of the reasoning path. Through empirical results and case study, we demonstrate that our proposed system is able to outperform several strong baselines on the {\dn} dataset. 



\end{document}